%% file: paper_final.tex
\documentclass[letterpaper, 10 pt, conference]{ieeeconf}  

\IEEEoverridecommandlockouts                              

\usepackage{preamble} 

\usepackage{flushend}

\title{\LARGE \bf Multi-Robot Local Motion Planning Using \\ Dynamic Optimization Fabrics
}

\author{Saray~Bakker*$^1$, Luzia~Knoedler*$^1$, Max~Spahn$^1$, Wendelin B\"{o}hmer$^2$ and Javier~Alonso-Mora$^1$
\thanks{*These authors contributed equally.}
\thanks{This paper has received funding from the European Union’s Horizon 2020 research and innovation programme under grant agreement No. 101017008, from the European Union through ERC, INTERACT, under Grant 101041863, and from Ahold Delhaize.
Views and opinions expressed are however those of the author(s) only and do not necessarily reflect those of the European
Union or the European Research Council Executive Agency.
Neither the European Union nor the granting authority can be
held responsible for them.}
\thanks{The authors are associated with the department of 1. Cognitive Robotics and
2. Electrical Engineering, Mathematics \& Computer Science,
        Delft University of Technology, The Netherlands,
    {\tt\small \{s.bakker-7, l.knoedler, m.spahn, j.w.bohmer j.alonsomora\}@tudelft.nl}}%
}

\begin{document}

\maketitle
\thispagestyle{empty}
\pagestyle{empty}
\input{commands.tex}

\begin{abstract}
In this paper, we address the problem of real-time motion planning for multiple robotic manipulators that operate in close proximity.
We build upon the concept of dynamic fabrics and extend them to multi-robot systems, referred to as Multi-Robot Dynamic Fabrics (MRDF). This geometric method enables a very high planning frequency for high-dimensional systems at the expense of being reactive and prone to deadlocks. To detect and resolve deadlocks, we propose Rollout Fabrics where MRDF are forward simulated in a decentralized manner. 
We validate the methods in simulated close-proximity pick-and-place scenarios with multiple manipulators, showing high success rates and real-time performance.

Code, video: \href{https://github.com/tud-amr/multi-robot-fabrics}{https://github.com/tud-amr/multi-robot-fabrics}

\end{abstract}

\input{sections_final/10_introduction}

\input{sections_final/20_related_work}

\input{sections_final/30_preliminaries}
\input{sections_final/35_multi_robot_fabrics}

\input{sections_final/40_method}
\input{sections_final/50_results}

\input{sections_final/60_conclusions}





\bibliographystyle{IEEEtran}
\bibliography{references_final}

\end{document}

%% file: commands.tex
\newcommand\X{\ensuremath{\mathcal{X}}}
\newcommand\Xr{\ensuremath{\mathcal{X}_{\textrm{rel}}}}
\newcommand\Xj{\ensuremath{\mathcal{X}_j}}
\newcommand\Q{\ensuremath{\mathcal{Q}}}
\newcommand\Rn{\ensuremath{\mathbb{R}^n}}
\newcommand\Rmj{\ensuremath{\mathbb{R}^{m_j}}}
\renewcommand\l{\ensuremath{\mathcal{L}}}
\renewcommand\le{\ensuremath{\mathcal{L}_e}}
\newcommand\ld{\ensuremath{\mathcal{L}_d}}
\renewcommand\lg{\ensuremath{\mathcal{L}_g}}
\newcommand\he{\ensuremath{\mathcal{H}_e}}
\newcommand\hd{\ensuremath{\mathcal{H}_d}}
\newcommand\dt{\ensuremath{\Delta t}}
\newcommand\Me{\ensuremath{\mat{M}_{\le}}}
\newcommand\fe{\ensuremath{\vec{f}_{\le}}}
\newcommand\Pe{\ensuremath{\mat{P}_{\le}}}
\newcommand\M{\ensuremath{\mat{M}}}
\newcommand\Mnh{\ensuremath{\mat{M}_{\textrm{nh}}}}
\newcommand\I{\ensuremath{\mat{I}}}
\newcommand\f{\ensuremath{\vec{f}}}
\newcommand\fnh{\ensuremath{\vec{f}_{\textrm{nh}}}}
\newcommand\h{\ensuremath{\vec{h}}}
\newcommand\zerovec{\ensuremath{\vec{0}}}
\newcommand\Spec{\ensuremath{\mathcal{S}}}
\newcommand\htwo{\ensuremath{\vec{h}_2}}
\newcommand\Md{\ensuremath{\mat{M}_d}}
\newcommand\fd{\ensuremath{\vec{f}_d}}
\newcommand\Mde{\ensuremath{\mat{M}_{de}}}
\newcommand\fde{\ensuremath{\vec{f}_{de}}}
\newcommand\forc{\ensuremath{\vec{\psi}}}
\newcommand\spec{\ensuremath{\left(\M,\f\right)_{\X}}}
\newcommand\x{\ensuremath{\vec{x}}}
\newcommand\xdot{\ensuremath{\dot{\x}}}
\newcommand\xddot{\ensuremath{\ddot{\x}}}
\newcommand\xj{\ensuremath{\vec{x}_j}}
\newcommand\xjdot{\ensuremath{\dot{\xj}}}
\newcommand\xjddot{\ensuremath{\ddot{\xj}}}
\newcommand\xb{\ensuremath{\bar{\vec{x}}}}
\newcommand\xbdot{\ensuremath{\dot{\xb}}}
\newcommand\xbddot{\ensuremath{\ddot{\xb}}}
\newcommand\xt{\ensuremath{\vec{\tilde{x}}}}
\newcommand\xtdot{\ensuremath{\dot{\xt}}}
\newcommand\xtddot{\ensuremath{\ddot{\xt}}}
\newcommand\xr{\ensuremath{\x_{\textrm{rel}}}}
\newcommand\xrdot{\ensuremath{\dot{\x}_{\textrm{rel}}}}
\newcommand\xrddot{\ensuremath{\ddot{\x}_{\textrm{rel}}}}
\newcommand\q{\ensuremath{\vec{q}}}
\newcommand\p{\ensuremath{\vec{p}}}
\newcommand\qdot{\ensuremath{\dot{\q}}}
\newcommand\qddot{\ensuremath{\ddot{\q}}}
\newcommand\qt{\ensuremath{\vec{\tilde{q}}}}
\newcommand\qtdot{\ensuremath{\dot{\qt}}}
\newcommand\qtddot{\ensuremath{\ddot{\qt}}}

\renewcommand\J{\ensuremath{\mat{J}_{\phi}}}
\newcommand\Jt{\ensuremath{\mat{J}^T_{\phi}}}
\newcommand\Jdot{\ensuremath{\dot{\mat{J}}_{\phi}}}
\newcommand\Jnh{\ensuremath{\mat{J}_{\textrm{nh}}}}
\newcommand\Jnht{\ensuremath{\mat{J}^T_{\textrm{nh}}}}
\newcommand\Jnhdot{\ensuremath{\dot{\mat{J}}_{\textrm{nh}}}}
\newcommand{\map}{\ensuremath{\phi}}
\newcommand{\mapt}{\ensuremath{\phi_t(\vec{q})}}
\newcommand{\mapd}{\ensuremath{\phi_d}}
\renewcommand{\g}{\ensuremath{\vec{g}(t)}}
\newcommand{\gp}{\ensuremath{\vec{g}^{\prime}(t)}}
\newcommand{\gpp}{\ensuremath{\vec{g}^{\prime\prime}(t)}}
\newcommand{\pull}[2]{\textrm{pull}_{#1}{#2}}
\newcommand{\energize}[2]{\textrm{energize}_{#1}{#2}}
\newcommand{\der}[2]{\partial_{#1}#2}
\newcommand{\dertwo}[2]{\partial^2_{#1}#2}
\newcommand{\derf}[2]{\frac{\partial#2}{\partial#1}}
\newcommand{\derftwo}[3]{\frac{\partial^2#3}{\partial#1\partial#2}}
\newcommand{\dert}[1]{\frac{d}{dt}#1}
\newcommand{\pinv}[1]{#1^{\dagger}}
\newcommand{\norm}[1]{\left\lVert#1\right\rVert}
\newcommand\mat[1]{\ensuremath{\bm{#1}}}
\renewcommand\vec[1]{\ensuremath{\bm{#1}}}
\renewcommand{\N}{\mathbb{N}}
\newcommand{\vq}{\mathbf{q}}
\newcommand{\R}{\mathbb{R}}

\acrodef{df}[DF]{Dynamic Fabrics}
\acrodef{sf}[SF]{Static Fabrics}
\acrodef{rf}[RF]{Rollout Fabrics}
\acrodef{srf}[SRF]{Static Rollout Fabrics}
\acrodef{drf}[DRF]{Dynamic Rollout Fabrics}
\acrodef{mrdf}[MRDF]{Multi-Robot Dynamic Fabrics}
\acrodef{mrsf}[MRSF]{Multi-Robot Static Fabrics}
\acrodef{mpc}[MPC]{Model Predictive Control}

%% file: sections_final/10_introduction.tex
\section{INTRODUCTION}
\label{sec:intro}
In several domains, such as manufacturing, medicine, and agriculture, it is common to have multiple manipulators operating in close proximity to each other. This arrangement improves efficiency and enables the successful execution of complex tasks \cite{feng2020overview}. However, when operating in dynamic environments, where conditions and obstacles can change, determining viable trajectories for multiple robots becomes challenging. 
Unlike static environments where all information can be pre-computed, dynamic environments require real-time planning to operate robots safely and efficiently. 

Multi-robot motion planning can be addressed using two main approaches: coupled and decoupled methods. Coupled approaches can provide optimal solutions, but they often suffer from high computational costs, especially when dealing with large numbers of robots with many degrees of freedom~(DOF)~\cite{christofides2013distributed}. On the other hand, decoupled approaches offer more scalability but cannot guarantee optimality. In dynamic environments, the ability to adapt online in real-time becomes essential.

\begin{figure}[t]
\centering
  \includegraphics[width=0.8\linewidth,clip]{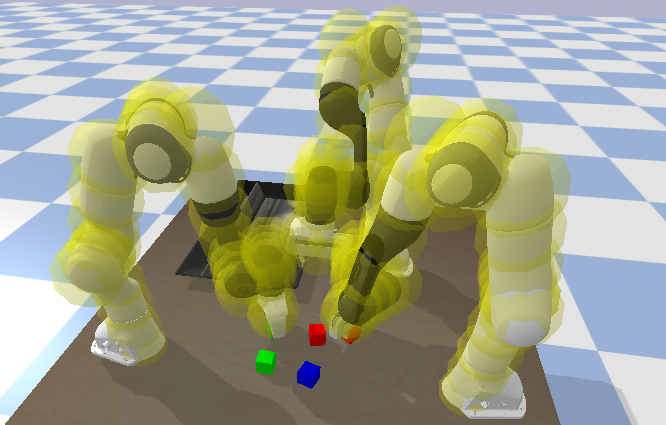}
  \caption{\small{Multi-robot pick-and-place scenario in close proximity. Franka Emika Pandas pick cubes avoiding collisions.}}
  \label{fig:exp_scenarios_3robots}
\end{figure}

However, even decoupled motion planning becomes challenging for complex systems.
While optimization-based approaches such as \ac{mpc} have been widely adopted for mobile robots, 
their high computational costs
make them less suitable for high-dimensional configuration spaces~\cite{spahn2021coupled}, \cite{edwards2021automatic},\cite{nubert2020safe}. Data-driven approaches can be utilized to speed up the optimization process \cite{nubert2020safe}, however, they often lack generalization capabilities and necessitate costly data acquisition processes \cite{hewing2020learning}.

In contrast, geometric approaches to local motion planning, such as Riemannian Motion
Policies~\cite{Ratliff2018, Cheng2020} and optimization fabrics~\cite{Ratliff2020}, offer superior scalability and thus high reactivity~\cite{spahn_dynamic_2023}.
With optimization fabrics, or fabrics for short, different components, such as collision avoidance and joint limit avoidance, are combined using Riemannian metrics, allowing for an iterative behavior design.
Because fabrics parse local motion planning into a differential equation of second order, the 
solution can be computed before runtime in a symbolic way~\cite{spahn_dynamic_2023}, 
thus saving computational costs during execution. This enables higher replanning frequencies making them advantageous for complex systems in dynamic environments. Furthermore, fabrics inherently promote asymptotic stability, which adds to their appeal as a framework for
reactive motion planning.

Fabrics, but also their predecessor Riemannian Motion Policies, have shown impressive results in several manipulator applications,
including dynamic and crowded environments~\cite{Wyk2022,Xie2020}.
Recently, fabrics have been generalized to dynamic scenarios, referred to as dynamic fabrics~\cite{spahn_dynamic_2023}.
As a drawback to highly reactive behavior, fabrics
are prone to local minima~\cite{spahn_dynamic_2023}, this is especially harmful in multi-robot scenarios where deadlocks are even more common than in single-robot applications.

This work focuses on the development of an online local motion planning algorithm to facilitate the simultaneous operation of multiple high DOF manipulators within a shared workspace. Our particular investigation revolves around the utilization of dynamic fabrics in multi-robot systems, specifically in the context of close-proximity pick-and-place scenarios.
To the best of the authors' knowledge, fabrics have not yet been employed within the context of multi-robot systems.
To tackle the challenges associated with close-proximity manipulation scenarios, we introduce a concept called \ac{rf}. This approach involves simulating the forward motion of multi-robot fabrics over a prediction horizon, enabling the detection of potential deadlocks. We propose a heuristic method to address the identified deadlocks, which involves adapting the goal-reaching parameters of the fabric formulation.
To this end, our contributions are as follows:
\begin{itemize}
    \item Extension of dynamic fabrics to multi-robot systems for high-DOF manipulators,
    referred to as Multi-Robot Dynamic Fabrics (MRDF),
    \item \ac{rf}, an approach to forward simulate MRDF enabling detection of future deadlocks or other undesired states,
    \item A heuristic approach for the resolution of deadlocks,
    \item Validation in simulation in close-proximity pick-and-place scenarios with high-DOF manipulators.
\end{itemize}

%% file: sections_final/20_related_work.tex
\section{RELATED WORK}\label{sec:related_work}
\subsection{Multi-Robot Motion Planning}\label{sec:related_work_multi_robot}
Regardless of the specific employed local motion planning algorithm, methods for multi-robot motion planning can be classified into two categories: coupled approaches and decoupled approaches. Coupled approaches compute the plans of all robots simultaneously and often provide guarantees regarding feasibility and optimality at the expense of a high computational burden.
Decoupled planners solve a sub-problem in isolation enabling rapid computation of feasible plans and relaxing communication requirements. 
However, they lack guarantees of completeness and optimality~\cite{solovey_finding_2016}.

Commonly used motion planning algorithms are sampling-based and optimization-based approaches, or combinations of both methods~\cite{hartmann_long-horizon_2023}. 
Sampling-based planners are well-suited for high-dimensional configuration spaces, making them a suitable choice for multi-robot systems. Rapidly exploring random trees~(RRT)~\cite{Kuffner2000} and the probabilistic roadmap method~(PRM)~\cite{kavraki_probabilistic_1996} can be extended to multi-robot systems by considering the robots as a single multi-arm robot or planning for each robot individually and adapting the velocities to avoid collisions~\cite{sanchez2002using}.
Further extensions improving efficiency are dRRT~\cite{solovey_finding_2016} which considers an implicit representation of a tensor product of roadmaps for the individual robots and its informed extension dRRT\textsuperscript{*}~\cite{shome_drrt_2020}.
Sampling-based methods are primarily used for static environments, where trajectories are initially planned for a specific task and then executed and thus mainly employed for offline motion planning.

Optimization-based planners~\cite{ratliff_chomp_2009} commonly take into account dynamic feasibility as well as static and dynamic collision avoidance as constraints.
Since an increasing number of constraints can lead to a significant increase in computational cost, it is crucial to efficiently implement constraints.
One commonly utilized optimization-based motion planner is \ac{mpc}~\cite{mayne_constrained_2000} which solves the optimization problem in a receding horizon manner, thereby continuously updating and optimizing the plan as new information becomes available.
For multi-robot motion planning, \ac{mpc} can be applied in a centralized~\cite{tika_optimal_2020} or distributed~\cite{tika_synchronous_2020, gafur_dynamic_2022} fashion. Since the centralized \ac{mpc} solves a coupled optimization problem, it has limited scalability~\cite{christofides2013distributed}. This motivates the development of distributed MPC formulations where each robot optimizes its own motion while considering the predicted behavior of the other robots.
For instance, \cite{gafur_dynamic_2022} present a distributed MPC approach formulated as a non-cooperative game that focuses on collision avoidance between multiple manipulators. However, their approach requires communication of the (extrapolated) predicted joint states of all neighbor robots at each time instance.
To avoid the costly solution of a constrained optimization problem, the planning
problem can be phrased as a purely geometric problem. Such methods fall into the field 
of geometric control. In a comparative study, it was shown that fabrics, a form of geometric control,
outperform \ac{mpc} in terms of collision avoidance and computation time for mobile manipulators~\cite{spahn_dynamic_2023}. 

Thus, in this work, we make use of the lightweight structure of optimization fabrics in a multi-robot setting. The fabrics formulation of the other robots only needs to be communicated once at the beginning of each task. Then at each time step, only the current configuration, velocity and goal information of the other robots must be communicated
among the robots. We further present approximations that can be used to reduce communication. 

\subsection{Geometric Control for Single-Robot Scenarios}
Operational space control, a control method applied to robotic systems, 
enables natural control of kinematically redundant
robots~\cite{Khatib1986,Khatib1987}. It was formalized in geometric control, which
utilizes differential geometry to achieve stable and converging
behavior~\cite{bullo2019geometric}. Recently,
Riemannian Motion Policies (RMP) introduced a trajectory generation method for
manipulation tasks, allowing composability by separating the importance metric and
forcing term~\cite{Ratliff2018,Cheng2020}.
Optimization fabrics were later introduced to fully decouple the
importance metrics and defining geometry, ensuring guaranteed convergence with
simple construction rules \cite{Ratliff2020,Xie2020,Li2021,meng2019neural}.
In \cite{spahn_dynamic_2023}, optimization fabrics are extended towards \ac{df} which incorporate path following and collision avoidance with moving obstacles. By proposing dynamic pullback operations, relative coordinate frames are explored introducing relative task positions, velocities, and accelerations. Convergence is guaranteed if the reference is bounded. Here we build upon optimization fabrics and \ac{df} and extend them to multi-robot environments.

%% file: sections_final/30_preliminaries.tex
\section{PRELIMINARIES}
In this section, we provide a concise introduction to the fundamental concepts necessary for trajectory generation using fabrics.
We first specify the required notations~\ref{sec:notations}, then we present spectral semi-sprays~\ref{sec:sprays} and their operations~\ref{sec:operations_sprays} which build the foundations for optimization fabrics~\ref{sec:optimization_fabrics} and \ac{df} \ref{sec:dynamic_fabrics}.
For a more detailed overview of fabrics and their theoretical foundations in differential geometry, we refer to~\cite{Ratliff2020, Wyk2022} and for \ac{df} to~\cite{spahn_dynamic_2023}. We further state the problem formulation for multi-robot fabrics~\ref{sec:problem_formulation}.

\subsection{Notations}\label{sec:notations}
We use $\q_t \in \mathcal{C}$ to denote the configuration of a robot at time $t$. Here, $\mathcal{C}$ is the robot's configuration space of dimension $n=DOF$. 
Similarly, $\dot{\q}_t$ and $\ddot{\q}_t$ define the corresponding instantaneous derivatives. 
To improve readability we will neglect the subscript $t$ in the following.

We define a set of $M$ task variables $\vec{x}_j \in \mathcal{X}_j$ for $j \in [M]$ with variable dimension $m_j \leq n$. We use the shorthand notation $[M]$ to denote $\{j \in \mathbb{N}: j \leq M \}$.
A differential map $\phi_j : \mathcal{C} 
\rightarrow \mathcal{X}_j$ with $\vec{x}_j = \phi_j(\q)$ relates the configuration space of a robot to the $j$-th task space. For instance, if a task variable is defined as the end-effector position, then
$\map_j$ is the positional part of the forward kinematics~(fk). If a task
variable is defined as the joint position, then $\map_j$ is the identity function. 
We assume that $\phi_j$ is smooth and twice differentiable and denote the map's Jacobian as 
$\vec{J}_{\phi_j} = \derf{\q}{\map_j}$
or $\vec{J}_{\phi_j} = \der{\q}{\map_j}$ for short. Thus, we can write the time derivatives of $\x_j$ as $\xdot_j = \vec{J}_{\phi_j}\qdot$ and $\xddot = \vec{J}_{\phi_j}\qddot + \dot{\vec{J}}_{\phi_j} \qdot$. In the following, the subscript $j$ is dropped to improve readability.

\subsection{Spectral Semi-Sprays}\label{sec:sprays}
\label{sub:semi_spectral_sprays}

Drawing inspiration from fundamental mechanics, optimization fabrics formulate motion policies as second-order dynamical systems, denoted by $\xddot = \pi(\x,\xdot)$ 
\cite{Cheng2020,Ratliff2020}. The motion policies are described by the differential equation $\M(\x,\xdot)\xddot + \f(\x,\xdot) = 0$, where the matrix $\M(\x,\xdot)$ is symmetric and invertible, and $\M(\x,\xdot)$ and $\f(\x,\xdot)$ depend on position and velocity variables. These second-order dynamical systems are referred to as \textit{spectral semi-sprays} or simply \textit{specs} $\Spec = \spec$. We drop the spec-subscript if the task manifold is clear from the context.

\subsection{Operations on Spectral Semi-Sprays}\label{sec:operations_sprays}
\label{sub:operations_on_specs}

The generation of complex trajectories involves various components, including collision avoidance and joint limits avoidance. 
In order to address these aspects, optimization fabrics utilize a weighted summation of metrics, enabling the integration of multiple components originating from different manifolds. The following operations, derived from operations on specs, play a crucial role in this process:

\textbf{Energization}: A spec can be \textit{energized} by incorporating Lagrangian energy, effectively equipping the spec with a metric. This process involves a spec of the form $\Spec_{\vec{h}} = (\mat{I},\vec{h})$ with identity matrix $\mat{I}$, a geometry defining term $\vec{h}$, and an energy Lagrangian \le{} with the derived equations of motion $\M_{\le}\xddot + \f_{\le} = 0$. The operation of \textit{energization} can be defined as follows:
\begin{equation}
\begin{split}
S_{\vec{h}}^{\le} &= \energize{\le}{S_{\vec{h}}} \\
&= (\Me, \fe + \Pe[\Me\vec{h} - \fe]),
\end{split}
\label{eq:energization}
\end{equation}
Here, $\Pe = \Me\left(\Me^{-1} - \frac{\xdot\xdot^T}{\xdot^T\Me\xdot}\right)$ represents an orthogonal projector. The resulting energy-conserving spec is referred to as an \textit{energized spec}, the operation itself is known as \textit{energization}. The energized system follows the same path and differs only by an acceleration along the direction of motion.

\textbf{Pullback}: Given a differential map $\map: \mathcal{C}\rightarrow\X$ and a spec \spec{}, the pullback is defined as
\begin{equation} \label{eq: pullback}
\pull{\map}{\spec} = {\left(\Jt\M\J, \Jt(\f+\Jdot\qdot)\right)}_{\mathcal{C}}.
\end{equation}
The pullback allows conversion between two distinct manifolds. For example, a spec defined in the robot's workspace can be pulled into the robot's configuration space using the pullback with $\map$ representing the forward kinematics.

\textbf{Summation}: For two specs, $\Spec_1 = {\left(\M_1,\f_1\right)}_{\mathcal{C}}$ and $\Spec_2 = {\left(\M_2,\f_2\right)}_{\mathcal{C}}$, their summation is defined as:
\begin{equation}
\Spec_1 + \Spec_2 = {\left(\M_1 + \M_2, \f_1 + \f_2\right)}_{\mathcal{C}}.
\label{eq:specs_summation}
\end{equation}
We denote the combined spec as $\tilde{\mathcal{S}} = (\tilde{\M}, \tilde{\f})_\mathcal{C}$.

By leveraging spectral semi-sprays and the aforementioned operations, joint limit avoidance, collision avoidance, or self-collision avoidance can be achieved.

\subsection{Optimization Fabrics} \label{sec:optimization_fabrics}

Above, we discussed the combination of tasks that can be used for combining different avoidance behaviors. Additionally, spectral semi-sprays can be influenced by a potential, where $\Spec_{\forc} = \left(\M,\f + \der{\x}{\forc}\right)$ is known as the \textit{forced spec}. The solution $\x(t)$ of the forced spec $\Spec_{\forc}$ converges to the minimum of the potential $\forc(\x)$ if constructed via the guidelines of optimization fabrics. 
The construction of optimization fabrics involves the following steps:

\noindent \textbf{1) Creation}: The initial spec representing an avoidance component is formulated in the form $\xddot + \vec{h}(\x,\xdot) = 0$. In this formulation, $\vec{h}$ is \textit{homogeneous of degree 2}, meaning that $\vec{h}(\x,\alpha\xdot) = \alpha^2\vec{h}(\x,\xdot)$.

\noindent \textbf{2) Energization}: The spec's geometry is energized using a Finsler structure~\cite[Definition 5.4]{Ratliff2020} through the energization operation in Eq.~\ref{eq:energization}. The combination of homogeneity of degree 2 and energization with the Finsler structure guarantees, as stated in~\cite[Theorem 4.29]{Ratliff2020}, that the energized spec forms a \textit{frictionless fabric}. A frictionless fabric is defined as optimizing the forcing potential \forc{} while being damped by a positive definite damping term~\cite[Definition 4.4]{Ratliff2020}. As a result, the trajectory converges to a local minima if damped.

\noindent \textbf{3) Combination}: 
All avoidance components are combined in the configuration space of the robot using the pullback and summation operations. Note that both operations are closed under the algebra designed by these operations. In other words, every pulled optimization fabric or the sum of two optimization fabrics is itself an optimization fabric.

\noindent \textbf{4) Forcing}: In the final step, the combined spec is forced by the potential \forc{} to the desired minimum with attractor weight $\gamma$ and damped with a positive definite constant damping matrix $\boldsymbol{B}$. This results in a system of the form 
\begin{equation} \label{eq:system}\tilde{\M}(\q, \qdot)\qddot + \tilde{\f}(\q, \qdot) + \gamma \der{\q}{\forc} + \boldsymbol{B}\qdot = 0.
\end{equation}

Solving Eq.~\eqref{eq:system} yields to the trajectory generation policy in acceleration form $\qddot = \tilde{\pi}(\q,\qdot)$. In the following, we will refer to optimization fabrics by \cite{Ratliff2020} as \ac{sf} since the formulation relies on the assumption that obstacles are static at each time step.

\subsection{Dynamic Fabrics}\label{sec:dynamic_fabrics}

Recently, \ac{df} have been proposed to handle dynamic environments, such as moving obstacles and reference path tracking~\cite{spahn_dynamic_2023}. By introducing a dynamic pull-back operation, 
relative coordinate systems, $\vec{x}_{rel} = \vec{x} - \bar{\vec{x}}$, can be exploited to integrate the velocity and acceleration of moving obstacles. This operation is defined as 
\begin{equation}
  \pull{\mapd}{{(\Md,\fd)}_{\Xr}} = {(\Md, \fd - \Md\ddot{\bar{\vec{x}}})}_{\X}, 
  \label{eq:dynamic_pull}
\end{equation}
where \Xr{} is the relative task manifold and $\ddot{\bar{\vec{x}}}$ is the acceleration of the moving obstacle. This dynamic pull-back
alongside the dynamic energization effectively renders the spec
$\left(\Md,\fd\right)_{\Xr}$ dependent on $\bar{\vec{x}}$ and $\dot{\bar{\vec{x}}}$. For clarity, we will refer to $(\bar{\vec{x}}, \dot{\bar{\vec{x}}}, \ddot{\bar{\vec{x}}})$ as $(\vec{x}_{obs}, \dot{\vec{x}}_{obs}, \ddot{\vec{x}}_{obs})$ to indicate the position, velocity, and acceleration of the moving obstacles. For simplicity, acceleration dependency is often disregarded.

\subsection{Multi-Robot Problem Formulation }\label{sec:problem_formulation}
In previous works, \ac{sf} and \ac{df} were presented for single-robot scenarios. Here, we formulate the multi-robot case as a decentralized collision avoidance problem. 
Consider a multi-robot scenario with $N$ robots with possibly different DOF, moving in close proximity in a shared workspace $\mathcal{W} \subseteq \mathbb{R}^3$. 
We introduce the superscript $i$ to refer to the $i$-th robot. 

\paragraph*{\textbf{Problem 1}}
 \label{prob:1}(\textit{Decentralized Multi-Robot Fabrics})

In the decentralized case, each robot~$i$ minimizes its own acceleration based on the states of the other robots~$\neg i$,
\begin{equation}\label{eq:policy}
\tilde{\pi}^i(\q^i, \qdot^i) = \qddot^i = \left(\tilde{\M}^{i}\right)^{-1}\left(\tilde{\f}^i+\gamma^i\der{\q^i}{\forc}^i + \beta\qdot^i\right),
\end{equation}

where $\tilde{\M}^i$ and $\tilde{\vec{f}}^i$ are dependent on the current configuration and velocity of the
robot as well as the states of the other robots $\neg i$.
In the following section, decentralized multi-robot fabrics are discussed in more detail and multi-robot collision avoidance is defined. 

%% file: sections_final/35_multi_robot_fabrics.tex
\section{Method}\label{sec:mf2}

In this section, we address multi-robot motion planning using fabrics as stated in ~\hyperref[prob:1]{Problem~1}. We first derive the collision avoidance formulation and present \ac{mrdf} in \cref{sec:mrca}. In \cref{sec:rollout_fabrics} Rollout Fabrics~(RF) are introduced to detect future deadlocks. Lastly, we present a heuristic approach to resolve the deadlocks in \cref{sec:resolving_deadlocks}. 

\subsection{Multi-Robot Dynamic Fabrics}\label{sec:mrca}
We first derive the collision-avoidance formulation  for \ac{mrdf} which requires
the configurations and velocities of the other robots. These can be determined using a perception pipeline or through communication.
Below, we only describe collision avoidance between robots, as the integration of static obstacles, dynamic obstacles, and joint limit avoidance were presented in \cite{spahn_dynamic_2023} and \cite{Ratliff2020}.

We approximate each robot using \textit{collision spheres}, see Fig.~\ref{fig:exp_scenarios_3robots},  with a center $\x^i_{obs,l}$ and radius $r^i_{obs,l}$ for $l \in [L^i]$ where $L^i$ is the number of collision spheres per robot. 
The \textit{collision obstacles} for robot $i$ are thus all collision spheres of the other robots, which is a total number of
$O^i~=~\sum_{p=1, p \neq i}^N L^p.
$
For each collision avoidance task $j \in [L^i  O^i]$ with relative task variable
$\vec{x}^i_{rel, j} = \vec{x}^i_{obs,l} - \vec{x}^p_{obs,l}$ for $p \neq i$, we consider a geometry $\ddot{\vec{x}}_{rel,j} + h(\vec{x}_{rel,j}, \dot{\vec{x}}_{rel,j}) = 0$. 

We derive the positions and velocities of the collision spheres using the forward kinematics and Jacobians, $\x_{obs,l}^p = \mathit{fk}_l(\q^p)$, $\dot{\x}_{obs,l}^p = J_l \qdot^p$.
Thus, Eq.~\eqref{eq:policy} becomes dependent on the state of all other robots resulting in 
\begin{equation}
  \label{eq:policy_dependent}
\tilde{\pi}^i(\q^i,\qdot^i, \q^{\neg i}, \qdot^{\neg i}, \vec{\theta}^i, \vec{\theta}^{\neg i}).  
\end{equation}
With a slight abuse of notation compared to Eq.~\eqref{eq:policy} we explicitly add the dependency of $\tilde{\pi}$ to parameter vectors $\vec{\theta}^i$ and $\vec{\theta}^{\neg i}$, e.g., the goal position $\vec{p}_\mathrm{goal}$ and attractor weight $\gamma$.

We refer to this equation as MRDF which is solved by every robot given that the configuration and velocity of the other robots are communicated, observed, or estimated.
In the following, we describe our proposed method to overcome deadlocks that
are likely to occur in this naive approach.

%% file: sections_final/40_method.tex
\subsection{Rollout Fabrics}\label{sec:rollout_fabrics}
In this section, we introduce the notion of \ac{rf}, 
an approach to forward propagate \ac{mrdf} over a prediction horizon enabling the detection of deadlocks by forward propagating \ac{mrdf}.
Since fabrics are a lightweight representation that can easily be communicated, each robot can transmit its symbolic fabrics policy $\tilde{\pi}^i$ in the beginning of the interaction.
We first assume that the current configuration and velocity, and goal configuration of all other robots are available at run-time. As mentioned before, this can be either observed or communicated. Later, we will relax this assumption.
Without loss of generality, we refer to the current time step as $k=0$. 
Each robot $i \in [N]$ propagates its own \ac{mrdf} and the \ac{mrdf} formulation of the other robots $\neg i$ forward over $K$ discrete steps covering the horizon $T=K\Delta t$.

To enhance clarity, we will now outline the approach from the perspective of the ego-robot $i$.
At each step $k$, the ego-robot $i$ computes its action $\ddot{\vec{q}}^{i}_k$ using Eq.~\ref{eq:policy_dependent}.
The actions of the other robots $\ddot{\vec{q}}^{\neg i}_k$ are derived using the communicated \ac{mrdf} formulations. 
Then, the configurations and velocities of each robot $i \in [N]$ at $k+1$ are computed given the configuration and velocities at step $k$, the action $\ddot{\vec{q}}^i_k$ and their respective goal configuration.
To avoid the need for simulating the complex dynamics of each robot, we approximate future configurations and velocities using a second-order integrator:
\begin{equation}\label{eq:integrator}
    \begin{bmatrix}
        \q_{k+1}^i \\ \qdot_{k+1}^i
    \end{bmatrix} = 
    \underbrace{\begin{bmatrix} 
    \vec{I}_n & \Delta t \vec{I}_n \\ 
    \vec{0}_n & \vec{I}_n
    \end{bmatrix}}_{\vec{A}}
    \begin{bmatrix}
        \q_{k}^i \\ \qdot_{k}^i
    \end{bmatrix} + 
    \underbrace{\begin{bmatrix}
        \vec{0}_n \\
        \Delta t \vec{I}_n
    \end{bmatrix}}_{\vec{B}}
    \qddot^i_k,
\end{equation}
with identity matrix $\vec{I}_{n}$ of size $n \times n$ and time-step $\Delta t$. The proposed approach is summarized in Algorithm~\ref{alg:1}.

Instead of communicating the goal configuration every time the goal has changed, it can constantly be estimated by assuming a constant velocity for the end effector,
\begin{equation}
\tilde{\p}_\mathrm{goal} = \x_\mathrm{ee} + H  \Delta t\vec{v}_\mathrm{ee},
\end{equation}
where $\x_\mathrm{ee}$ and $\vec{v}_\mathrm{ee}$ are the end effector position and velocity, respectively. The integer scaling factor $H$ determines the duration for which the constant velocity model predicts the future goal.
\ac{rf} considering the estimated goal $\tilde{\p}_\mathrm{goal}$ are in the following referred to as RF-CV.

The rollouts provide us with a set of configuration, velocity, and acceleration estimates along $T$, which can be used to detect deadlocks.
We apply a heuristic that detects deadlocks where at least two robots have average velocities along the prediction horizon $\bar{v}^i = \frac{1}{K}\sum^K_{k=0} \lVert \qdot^i_k \rVert_2$ below the threshold $v_{d, min}$. An additional condition for a deadlock is that the position of the end-effectors $\vec{x}_{ee}$ of the robots have to be within a distance $d_{ee, c}$ from each other. Thus, a deadlock is detected if the following condition holds for any $p \neq i$:
\begin{equation} \label{eq:detect_deadlock}
    \resizebox{0.9\linewidth}{!}{
     $\left(  \bar{v}^i < v_{\mathrm{d}, \mathrm{min}} \wedge  \bar{v}^p < v_{d, min} \right)
    \wedge \left( \lVert \vec{x}_{\mathrm{ee}}^i - \vec{x}_{\mathrm{ee}}^p \rVert_2 < d_{\mathrm{ee}, \mathrm{c}} \right).
    $
    }
\end{equation}
If no deadlock is present, the ego-robot's action is the acceleration $\ddot{\vec{q}}^i_0$.
A deadlock is assumed to be resolved when the predicted velocities are above the velocity minimum, $v_{d, min}$ and a time of $t_{d, min}$ has passed, or when one of the robots has reached its goal. The time requirement is introduced to avoid the robots from switching between normal and deadlock-resolving behavior.

Since we can approximate the current configurations and velocities of the other robots with \ac{rf}, it is not required for each robot to communicate this at each time step. Instead, these can be communicated at a lower frequency and approximated using forward propagation if no current information is available.
\begin{algorithm}
    \small
    \caption{Rollout Fabrics}
    \renewcommand{\algorithmicrequire}{\textbf{Input:}}
    \begin{algorithmic}
    \REQUIRE $(\q^i_0, \qdot^i_0) = (\q^i_t, \qdot^i_t), \ \ \forall i\in [N]$
    \hfill $\triangleright$\textit{Initialization}\\
    \FOR{$k=0:K$}
        \STATE $\vec{x}_{\mathrm{obst}}^i = \mathit{fk}(\q^i_k, \qdot^i_k), \ \ \forall i \in [N]$ 
        \hfill $\triangleright$\textit{ obstacle positions}
        \STATE $\dot{\vec{x}}_{\mathrm{obst}}^i = J_l(\q^i_k, \qdot^i_k), \ \ \forall i \in [N]$ 
        \hfill $\triangleright$\textit{ obstacle velocities}
        \STATE $\qddot_{k}^i = \tilde{\pi}^i(\q^i,\qdot^i, \q^{\neg i}, \qdot^{\neg i}, \vec{\theta}^i, \vec{\theta}^{\neg i}), \ \  \forall i \in [N]$
        \hfill $\triangleright$\textit{ \ac{mrdf}}
        \STATE$\begin{bmatrix} \q_{k+1}^i \\ \qdot_{k+1}^i \\
        \end{bmatrix} = \vec{A} \begin{bmatrix} \q_{k}^i \\ \qdot_{k}^i 
        \end{bmatrix} + \vec{B} \qddot^i_k, \ \ \forall i \in [N]$
        \hfill $\triangleright$\textit{ Model}
    \ENDFOR
    \STATE $deadlock = $ Eq.~\ref{eq:detect_deadlock}
    \hfill $\triangleright$\textit{ Detect deadlock}
    \end{algorithmic}
    \label{alg:1}
\end{algorithm}
\vspace{-5mm}

\subsection{Resolving Deadlocks}\label{sec:resolving_deadlocks}
Here, we describe how the detected deadlocks can be addressed using a heuristic approach, which is summarized in Algorithm~\ref{alg:2} for the two-robot case.
If a deadlock is detected, a hierarchy is defined and communicated, based on the proximity of the robots to their respective goal or at random for a perfectly symmetric scenario. With this method, we do not claim completeness to resolve all possible deadlocks. 
Note that if all robots apply the same heuristic and the configurations and desired goals of the other robots are known, the hierarchy does not need to be communicated.

To resolve a deadlock, the lower-priority robot's goal is set opposite to the higher-prioritized robot's goal, $goal\_low()$. Additionally, the weight for goal-reaching $\gamma$ of the higher-prioritized robot is increased to $\gamma_{high}$. 

\begin{algorithm}
    \small
    \caption{Resolving deadlocks}
    \begin{algorithmic}
        \IF{deadlock}
        \STATE $g_{\mathrm{low\_priority}} = goal\_low(\q)$,
        \hfill $\triangleright$\textit{ Change goal}
        \STATE $\gamma_{\mathrm{priority}} = \gamma_{\mathrm{high}}$,
        \hfill $\triangleright$\textit{ Change weight}
        \STATE $\qddot_t = \tilde{\pi}(\q^i_t,\qdot^i_t, \q^{\neg i}_t,\qdot^{\neg i}_t, \vec{\theta}^i, \vec{\theta}^{\neg i}), \ \  \forall i \in [N]$
        \hfill $\triangleright$\textit{ \ac{mrdf}}
        \ELSE
        \STATE $\qddot_t = \qddot_{0}$,
        \hfill $\triangleright$\textit{ Apply first action}
        \ENDIF
    \end{algorithmic}
    \label{alg:2}
\end{algorithm}

%% file: sections_final/50_results.tex
\section{RESULTS}\label{sec:results}

\begin{table}[b]
\renewcommand{\arraystretch}{1}
\caption{Parameters}\label{hp}
\centering
\begin{tabular}{|p{3.3cm}|p{1.1cm}||p{1.8cm}|p{0.6cm}|}
 \hline
  Max time $T_{max}$ & \SI{70}{\second}  & Time step $\Delta t$  & \SI{0.01}{\second}\\ 
 \hline
   \# of collision spheres $L^i$ & \SI{32}{} & Pred. steps $K$ & \SI{10}{}\\
 \hline
  Min vel deadlock $v_{d, min}$& \SI{0.03}{rad/\second} & Radius $r_{obst}$& \SI{8}{cm} \\
  \hline 
  Min time deadlock $t_{d, min}$ & \SI{3.0}{\second} &  Weight $\gamma_{high}$ & \SI{3}{}\\
  \hline 
 Min distance deadlock $d_{ee,c}$ & \SI{0.35}{m} & Weight $\gamma_{low}$ & \SI{2}{}\\
 \hline
\end{tabular}
\end{table}
Here, we assess the performance of \ac{rf} with deadlock resolution heuristics and compare them against \ac{mrdf}.
For a comparison between fabrics and MPC, we refer to~\cite{spahn_dynamic_2023}.

\begin{figure*}[ht]   
\subfloat[Initial pose]{\includegraphics[width=0.19\textwidth, keepaspectratio]{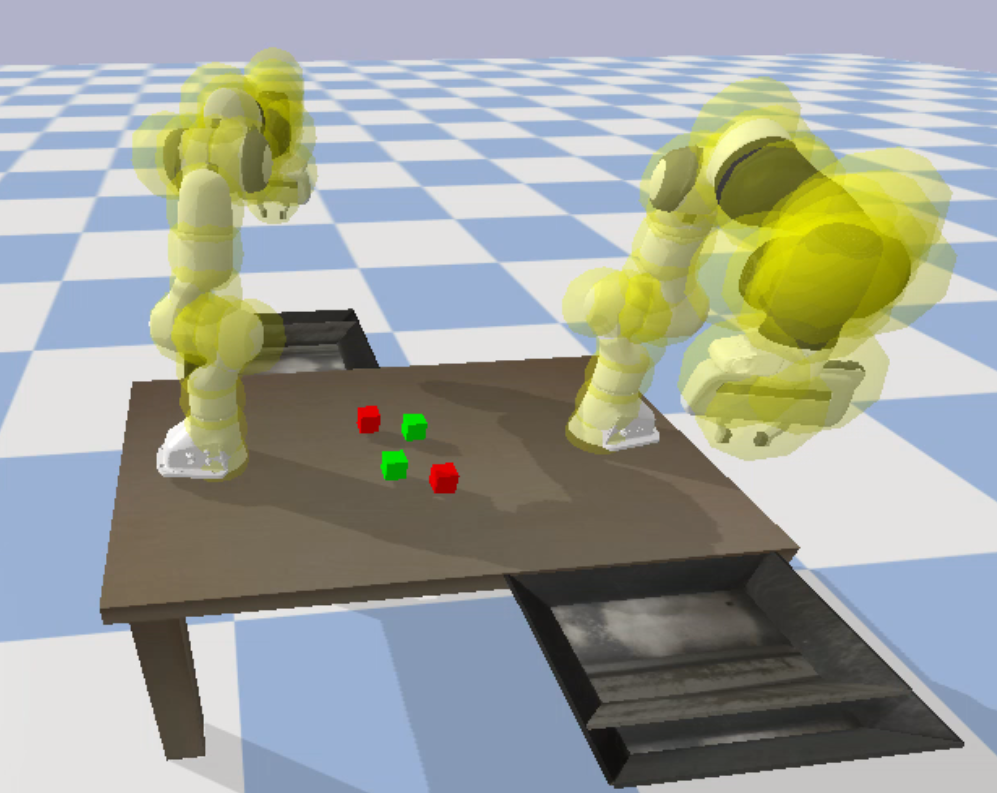}}\label{fig:subfig5}
\hspace{\fill}
\subfloat[Deadlock detected]{\includegraphics[width=0.19\textwidth, keepaspectratio]{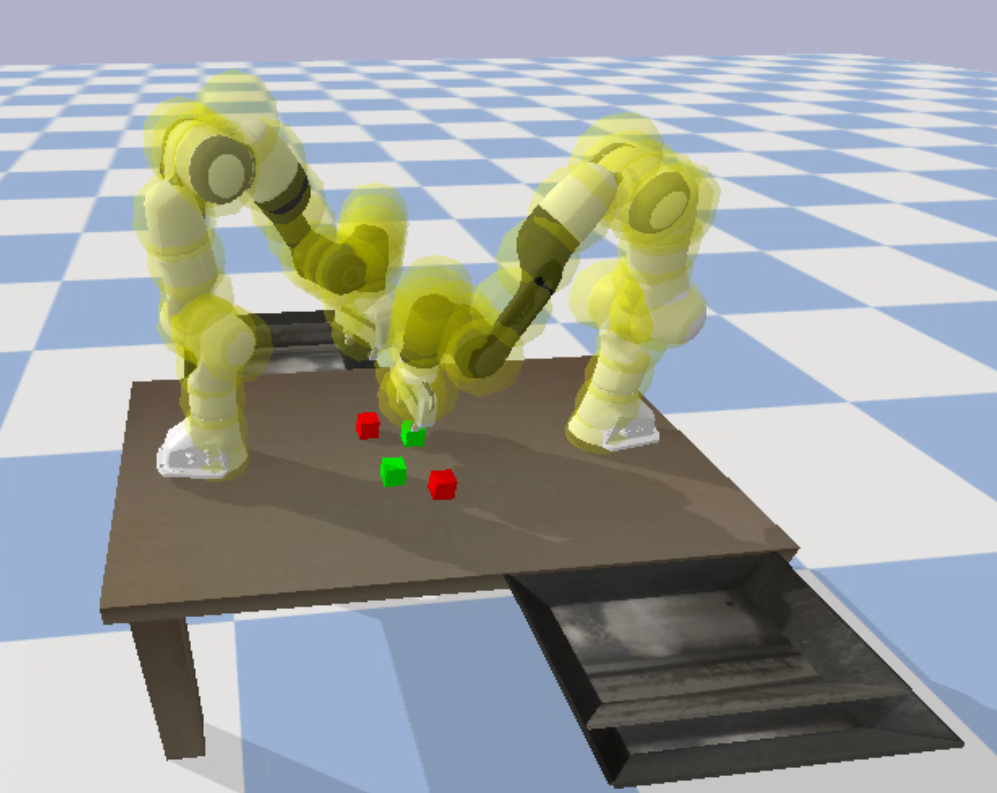}}\label{fig:subfig6}
\hspace{\fill}
\subfloat[Deadlock resolving]{\includegraphics[width=0.19\textwidth, keepaspectratio]{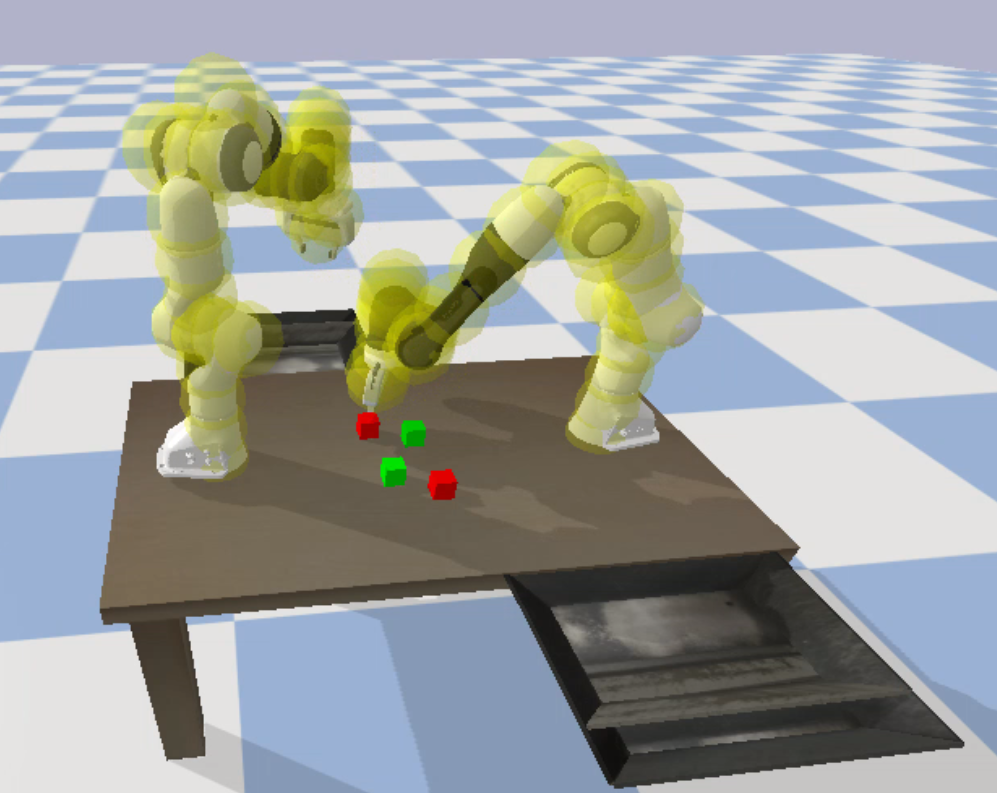}}\label{fig:subfig7}
\hspace{\fill}
\subfloat[Deadlock resolved]{\includegraphics[width=0.19\textwidth, keepaspectratio]{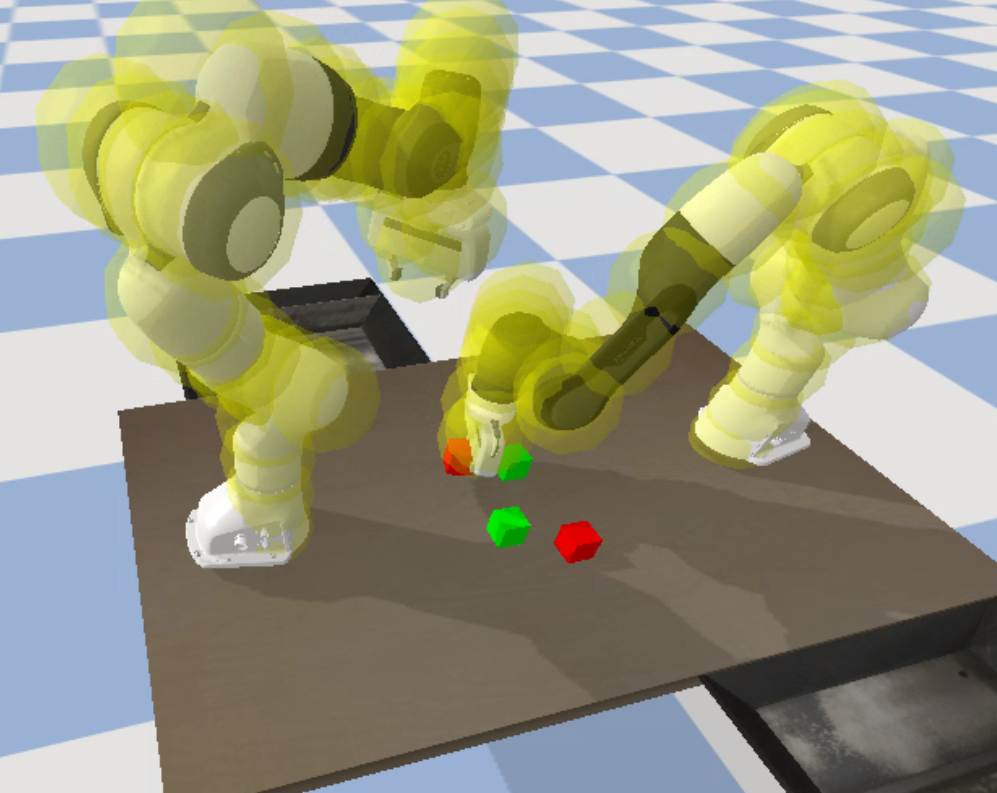}}\label{fig:subfig8}
\hspace{\fill}
\subfloat[Block delivered]{\includegraphics[width=0.19\textwidth, keepaspectratio]{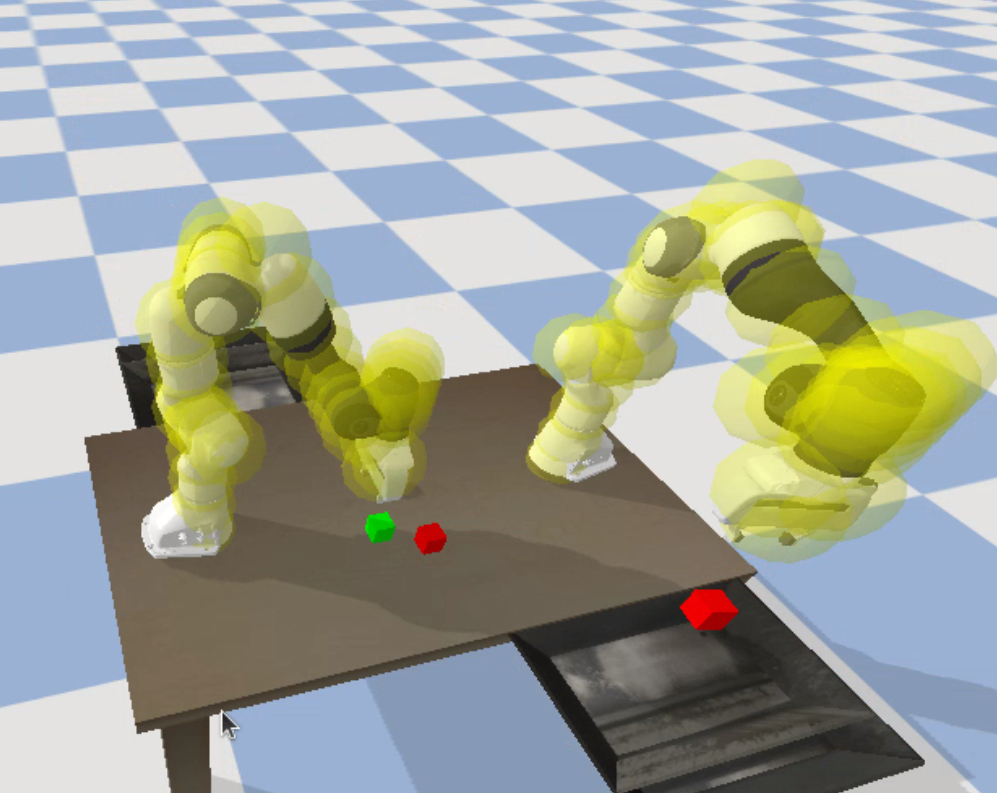}}\label{fig:subfig9}
\caption[]{\small{Selected time frames of \ac{rf} resolving deadlocks}}
\label{fig:screenshots_deadlock}
\vspace{-4mm}
\end{figure*}

\begin{table*} [bt]
	\begin{center}
 	\caption{\small{Statistics for 50 scenarios of our proposed methods \ac{rf} and RF-CV compared to 50 scenarios of \ac{mrdf}.}}
    \label{tab: results_simulation}
		\begin{tabular}{c|c|c|c|c|c}
			  & Success-Rate & Time-to-Success [s] & Collision-rate & Min Clearance [m] & Computation Time [ms] \\
			\hline
			\ac{mrdf}    & 0.73 $\pm$ 0.40        & 36.4 $\pm$ 15.8   & 0.31     & 0.004 $\pm$ 0.004     & 4 $\pm$ 0.9\\
			\ac{rf}      & 0.97 $\pm$ 0.11       & 25.0 $\pm$ 4.0    & 0.04   & 0.022 $\pm$ 0.013     & 20.5 $\pm$ 0.8\\
            RF-CV        & 0.98 $\pm$ 0.09      & 24.7 $\pm$ 2.8      & 0.04      & 0.024 $\pm$ 0.015       & 29 $\pm$ 1.0
		\end{tabular}
	\end{center}
    \vspace{-1cm}
\end{table*}

\subsection{Experimental Setup and Performance Metrics}
We consider multi-robot close-proximity pick-and-place scenarios as illustrated in \Cref{fig:exp_scenarios_3robots}~and~\ref{fig:screenshots_deadlock} and the video. Specifically, we apply multiple $7$~DOF Franka Emika Pandas that are tasked to pick up their assigned cubes from a table and place them in a tray. Simulations are performed using the Pybullet physics simulation \cite{coumans2019pybullet} and the open source toolbox \href{https://github.com/maxspahn/gym_envs_urdf}{urdfenvs}. In the experiments, besides multi-robot collision avoidance, collision avoidance between each robot and the table is considered, as well as joint limit avoidance.
The robots each have to pick and place two cubes. 
To evaluate the performance we consider the below metrics for a two-robot scenario with randomized initial cube positions:
\begin{itemize}
    \item \textit{Success Rate}: The ratio of successfully grasped and placed cubes within the time window $T_{max}$ over the total number of cubes. 
    \item \textit{Time-to-Success}: Time to complete the pick-and-place scenario for all robots successfully. 
    \item \textit{Collision Rate}: Ratio of scenarios where at least one collision has occurred over all runs. A collision is registered if any collision sphere of a robot collides with either the environment or the collision spheres of the other robot.
    \item \textit{Minimum Clearance}: Minimum distance between the collision spheres of the robots.
    \item \textit{Computation Time}: The time to compute an action via the local motion planner at each time step.
\end{itemize}
Note, the above performance metrics, excluding the success rate and computation time, are only evaluated if the concerned motion planner succeeds.
We execute our planner on a standard laptop (i7-12700H) without parallelization. 
All parameters are summarized in \Cref{hp}. Joint velocities are provided as inputs to the robots by integrating $\vec{\ddot{q}}^i, \ \forall i \in [N]$.

\begin{figure}
    \centering    \includegraphics[width=0.94\columnwidth]{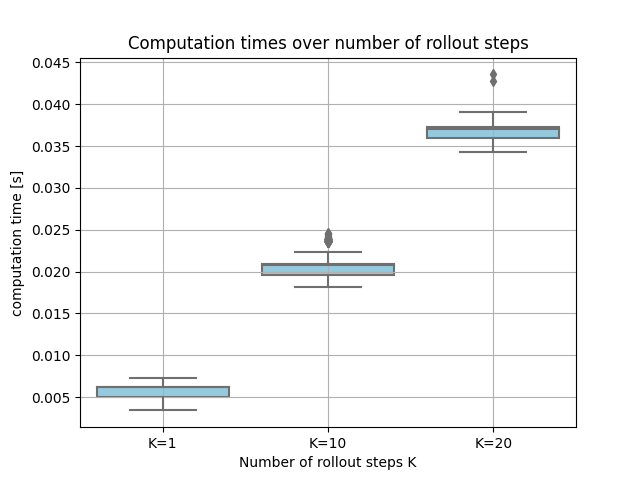}
    \caption{\small{Scaling of computation times with horizon length $K$.}}
    \label{fig:solver_times}
\end{figure}
\begin{figure} [bt]
    \centering    \includegraphics[width=0.94\columnwidth]{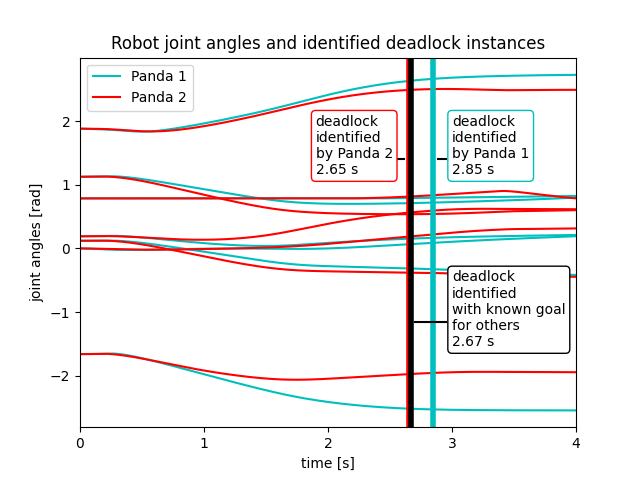}
    \caption{\small{Robot joint angles and identified deadlock instances. Panda 1 and Panda 2 apply Rollout Fabrics with an estimated goal for the other robot, respectively. Furthermore, the deadlock identified with full knowledge of the other robot's goal is displayed.}}
    \label{fig:deadlock_angles}
\end{figure}

\subsection{Simulation Experiments}
Table~\ref{tab: results_simulation} displays the performance across 50 scenarios with randomized cube positions for MRDF, RF, and RF-CV. The benefits of \ac{rf} combined with a heuristic approach to resolve deadlocks are clearly illustrated by a success rates of 97\% compared to 73\% for \ac{mrdf}.
The minimum clearance and collision-rate are also improved for \ac{rf} compared to \ac{mrdf}, since the deadlocks in \ac{mrdf} cause close-to-collision configurations.
Due to the rollouts being efficiently implemented using a symbolic Casadi function \cite{Andersson2019} defined beforehand and called during operation, the computation time of 20.5~ms is realistic for real-time local motion planning. 
Although a longer prediction horizon results in earlier detection of a deadlock, it comes with the drawback of increased computation times, see Fig.~\ref{fig:solver_times}. 

While RF assume that the goal configurations of the other robots are known, RF-CV removes this assumption by estimating the goal of the other robots. Figure~\ref{fig:deadlock_angles} illustrates the joint angles of two robots that have reached a deadlock according to Eq.~\eqref{eq:detect_deadlock}. For this specific scenario, the robots detect the deadlock with a time difference of 0.2~s when using goal estimation and deviate from RF with known goals. 
The success-rate, minimum clearance, collision-rate, and time-to-success for RF-CV are similar to \ac{rf}, as can be seen in Table~\ref{tab: results_simulation}. 
Applying RF-CV improved collision avoidance when compared to MRDF without requiring additional communication.
Therefore, the goal estimation is beneficial in set-ups where communication is challenging or unreliable.

%% file: sections_final/60_conclusions.tex
\section{CONCLUSIONS}
In this work, we showcased the applicability of dynamic fabrics to multi-robot scenarios. 
Additionally, dynamic fabrics are extended to \acf{rf}, where dynamic fabrics are propagated forward in time. These future predictions are used to detect and resolve deadlock scenarios.
Simulation experiments are performed with multiple manipulators where each robot performs a pick-and-place task in close proximity to the other robots. 
Analyzing the scenarios with two manipulators, the success rate of the proposed \ac{rf} in combination with the proposed deadlock resolution is increased compared to multi-robot dynamic fabrics. Additionally, we analyze \ac{rf} with goal estimation which removes the requirement for communicating updated goals.
For future research, we want to explore the applicability of \ac{rf} to human-centered environments and analyze how the strategy of rollouts can be used for manipulation tasks involving physical interaction. Our approach would thereby benefit from advances in environment and robot representation.
Furthermore, we intend to show the applicability of \ac{rf} in real-world experiments.

%% file: paper_final.bbl
\begin{thebibliography}{10}
\providecommand{\url}[1]{#1}
\csname url@samestyle\endcsname
\providecommand{\newblock}{\relax}
\providecommand{\bibinfo}[2]{#2}
\providecommand{\BIBentrySTDinterwordspacing}{\spaceskip=0pt\relax}
\providecommand{\BIBentryALTinterwordstretchfactor}{4}
\providecommand{\BIBentryALTinterwordspacing}{\spaceskip=\fontdimen2\font plus
\BIBentryALTinterwordstretchfactor\fontdimen3\font minus \fontdimen4\font\relax}
\providecommand{\BIBforeignlanguage}[2]{{%
\expandafter\ifx\csname l@#1\endcsname\relax
\typeout{** WARNING: IEEEtran.bst: No hyphenation pattern has been}%
\typeout{** loaded for the language `#1'. Using the pattern for}%
\typeout{** the default language instead.}%
\else
\language=\csname l@#1\endcsname
\fi
#2}}
\providecommand{\BIBdecl}{\relax}
\BIBdecl

\bibitem{feng2020overview}
Z.~Feng, G.~Hu, Y.~Sun, and J.~Soon, ``An overview of collaborative robotic manipulation in multi-robot systems,'' \emph{Annual Reviews in Control}, vol.~49, pp. 113--127, 2020.

\bibitem{christofides2013distributed}
P.~D. Christofides, R.~Scattolini, D.~M. de~la Pena, and J.~Liu, ``Distributed model predictive control: A tutorial review and future research directions,'' \emph{Computers \& Chemical Engineering}, vol.~51, pp. 21--41, 2013.

\bibitem{spahn2021coupled}
M.~Spahn, B.~Brito, and J.~Alonso-Mora, ``Coupled mobile manipulation via trajectory optimization with free space decomposition,'' in \emph{2021 IEEE International Conference on Robotics and Automation (ICRA)}.\hskip 1em plus 0.5em minus 0.4em\relax IEEE, 2021, pp. 12\,759--12\,765.

\bibitem{edwards2021automatic}
W.~Edwards, G.~Tang, G.~Mamakoukas, T.~Murphey, and K.~Hauser, ``Automatic tuning for data-driven model predictive control,'' in \emph{2021 IEEE International Conference on Robotics and Automation (ICRA)}.\hskip 1em plus 0.5em minus 0.4em\relax IEEE, 2021, pp. 7379--7385.

\bibitem{nubert2020safe}
J.~Nubert, J.~K{\"o}hler, V.~Berenz, F.~Allg{\"o}wer, and S.~Trimpe, ``Safe and fast tracking on a robot manipulator: Robust mpc and neural network control,'' \emph{IEEE Robotics and Automation Letters}, vol.~5, no.~2, pp. 3050--3057, 2020.

\bibitem{hewing2020learning}
L.~Hewing, K.~P. Wabersich, M.~Menner, and M.~N. Zeilinger, ``Learning-based model predictive control: Toward safe learning in control,'' \emph{Annual Review of Control, Robotics, and Autonomous Systems}, vol.~3, pp. 269--296, 2020.

\bibitem{Ratliff2018}
N.~D. Ratliff, J.~Issac, D.~Kappler, S.~Birchfield, and D.~Fox, ``Riemannian motion policies,'' \emph{arXiv preprint arXiv:1801.02854}, 2018.

\bibitem{Cheng2020}
C.-A. Cheng, M.~Mukadam, J.~Issac, S.~Birchfield, D.~Fox, B.~Boots, and N.~Ratliff, ``Rmp flow: A computational graph for automatic motion policy generation,'' in \emph{Algorithmic Foundations of Robotics XIII: Proceedings of the 13th Workshop on the Algorithmic Foundations of Robotics 13}.\hskip 1em plus 0.5em minus 0.4em\relax Springer, 2020, pp. 441--457.

\bibitem{Ratliff2020}
N.~D. Ratliff, K.~Van~Wyk, M.~Xie, A.~Li, and M.~A. Rana, ``Optimization fabrics,'' \emph{arXiv preprint arXiv:2008.02399}, 2020.

\bibitem{spahn_dynamic_2023}
M.~Spahn, M.~Wisse, and J.~Alonso-Mora, ``Dynamic {Optimization} {Fabrics} for {Motion} {Generation},'' \emph{IEEE Transactions on Robotics}, pp. 1--16, 2023.

\bibitem{Wyk2022}
K.~Van~Wyk, M.~Xie, A.~Li, M.~A. Rana, B.~Babich, B.~Peele, Q.~Wan, I.~Akinola, B.~Sundaralingam, D.~Fox \emph{et~al.}, ``Geometric fabrics: Generalizing classical mechanics to capture the physics of behavior,'' \emph{IEEE Robotics and Automation Letters}, vol.~7, no.~2, pp. 3202--3209, 2022.

\bibitem{Xie2020}
M.~Xie, K.~Van~Wyk, A.~Li, M.~A. Rana, Q.~Wan, D.~Fox, B.~Boots, and N.~Ratliff, ``Geometric fabrics for the acceleration-based design of robotic motion,'' \emph{arXiv preprint arXiv:2010.14750}, 2020.

\bibitem{solovey_finding_2016}
K.~Solovey, O.~Salzman, and D.~Halperin, ``Finding a needle in an exponential haystack: Discrete rrt for exploration of implicit roadmaps in multi-robot motion planning,'' \emph{The International Journal of Robotics Research}, vol.~35, no.~5, pp. 501--513, 2016.

\bibitem{hartmann_long-horizon_2023}
V.~N. Hartmann, A.~Orthey, D.~Driess, O.~S. Oguz, and M.~Toussaint, ``Long-horizon multi-robot rearrangement planning for construction assembly,'' \emph{IEEE Transactions on Robotics}, vol.~39, no.~1, pp. 239--252, 2022.

\bibitem{Kuffner2000}
J.~J. Kuffner and S.~M. LaValle, ``Rrt-connect: An efficient approach to single-query path planning,'' in \emph{2000 IEEE International Conference on Robotics and Automation (ICRA)}, vol.~2.\hskip 1em plus 0.5em minus 0.4em\relax IEEE, 2000, pp. 995--1001.

\bibitem{kavraki_probabilistic_1996}
L.~E. Kavraki, P.~Svestka, J.-C. Latombe, and M.~H. Overmars, ``Probabilistic roadmaps for path planning in high-dimensional configuration spaces,'' \emph{IEEE Transactions on Robotics and Automation}, vol.~12, no.~4, pp. 566--580, 1996.

\bibitem{sanchez2002using}
G.~Sanchez and J.-C. Latombe, ``Using a prm planner to compare centralized and decoupled planning for multi-robot systems,'' in \emph{2002 IEEE International Conference on Robotics and Automation (ICRA)}, vol.~2.\hskip 1em plus 0.5em minus 0.4em\relax IEEE, 2002, pp. 2112--2119.

\bibitem{shome_drrt_2020}
R.~Shome, K.~Solovey, A.~Dobson, D.~Halperin, and K.~E. Bekris, ``drrt*: Scalable and informed asymptotically-optimal multi-robot motion planning,'' \emph{Autonomous Robots}, vol.~44, no. 3-4, pp. 443--467, 2020.

\bibitem{ratliff_chomp_2009}
N.~Ratliff, M.~Zucker, J.~A. Bagnell, and S.~Srinivasa, ``Chomp: Gradient optimization techniques for efficient motion planning,'' in \emph{2009 IEEE international conference on robotics and automation}.\hskip 1em plus 0.5em minus 0.4em\relax IEEE, 2009, pp. 489--494.

\bibitem{mayne_constrained_2000}
D.~Q. Mayne, J.~B. Rawlings, C.~V. Rao, and P.~O. Scokaert, ``Constrained model predictive control: Stability and optimality,'' \emph{Automatica}, vol.~36, no.~6, pp. 789--814, 2000.

\bibitem{tika_optimal_2020}
A.~Tika, N.~Gafur, V.~Yfantis, and N.~Bajcinca, ``Optimal scheduling and model predictive control for trajectory planning of cooperative robot manipulators,'' \emph{IFAC-PapersOnLine}, vol.~53, no.~2, pp. 9080--9086, 2020.

\bibitem{tika_synchronous_2020}
A.~Tika and N.~Bajcinca, ``Synchronous minimum-time cooperative manipulation using distributed model predictive control,'' in \emph{2020 IEEE/RSJ International Conference on Intelligent Robots and Systems (IROS)}.\hskip 1em plus 0.5em minus 0.4em\relax IEEE, 2020, pp. 7675--7681.

\bibitem{gafur_dynamic_2022}
N.~Gafur, G.~Kanagalingam, and M.~Ruskowski, ``Dynamic collision avoidance for multiple robotic manipulators based on a non-cooperative multi-agent game,'' \emph{arXiv preprint arXiv:2103.00583}, 2021.

\bibitem{Khatib1986}
O.~Khatib, ``Real-time obstacle avoidance for manipulators and mobile robots,'' \emph{The international journal of robotics research}, vol.~5, no.~1, pp. 90--98, 1986.

\bibitem{Khatib1987}
------, ``A unified approach for motion and force control of robot manipulators: The operational space formulation,'' \emph{IEEE Journal on Robotics and Automation}, vol.~3, no.~1, pp. 43--53, 1987.

\bibitem{bullo2019geometric}
F.~Bullo and A.~D. Lewis, \emph{Geometric control of mechanical systems: modeling, analysis, and design for simple mechanical control systems}.\hskip 1em plus 0.5em minus 0.4em\relax Springer, 2019, vol.~49.

\bibitem{Li2021}
A.~Li, C.-A. Cheng, M.~A. Rana, M.~Xie, K.~Van~Wyk, N.~Ratliff, and B.~Boots, ``Rmp2: A structured composable policy class for robot learning,'' \emph{arXiv preprint arXiv:2103.05922}, 2021.

\bibitem{meng2019neural}
X.~Meng, N.~Ratliff, Y.~Xiang, and D.~Fox, ``Neural autonomous navigation with riemannian motion policy,'' in \emph{2019 International Conference on Robotics and Automation (ICRA)}.\hskip 1em plus 0.5em minus 0.4em\relax IEEE, 2019, pp. 8860--8866.

\bibitem{coumans2019pybullet}
E.~Coumans and Y.~Bai, ``Pybullet, a python module for physics simulation for games, robotics and machine learning,'' \url{http://pybullet.org}, 2016.

\bibitem{Andersson2019}
J.~A. Andersson, J.~Gillis, G.~Horn, J.~B. Rawlings, and M.~Diehl, ``Casadi: a software framework for nonlinear optimization and optimal control,'' \emph{Mathematical Programming Computation}, vol.~11, pp. 1--36, 2019.

\end{thebibliography}
